\documentclass{article} %
\usepackage{iclr2024_conference,times}

\usepackage{amsmath,amsfonts,bm}

\def\eqref#1{equation~\ref{#1}}

\def\1{\bm{1}}

\DeclareMathAlphabet{\mathsfit}{\encodingdefault}{\sfdefault}{m}{sl}
\SetMathAlphabet{\mathsfit}{bold}{\encodingdefault}{\sfdefault}{bx}{n}

\usepackage{hyperref}
\usepackage{url}
\usepackage{booktabs,graphicx,colortbl}

\definecolor{citecolor}{HTML}{0071BC}
\definecolor{linkcolor}{HTML}{ED1C24}
\definecolor{Gray}{gray}{0.86}

\hypersetup{colorlinks=True, citecolor=citecolor, urlcolor=magenta}

\title{A Change Detection Reality Check}

\iclrfinalcopy

\author{Isaac Corley\\
University of Texas at San Antonio\\
San Antonio, TX, USA \\
\texttt{isaac.corley@utsa.edu} \\
\And
Caleb Robinson \& Anthony Ortiz \\
Microsoft AI for Good Research Lab \\
Redmond, WA, USA \\
\texttt{\{caleb.robinson,anthony.ortiz\}@microsoft.com}
}

\begin{document}

\maketitle

\begin{abstract}
In recent years, there has been an explosion of proposed change detection deep learning architectures in the remote sensing literature. These approaches claim to offer state-of-the-art performance on different standard benchmark datasets. However, has the field truly made significant progress? In this paper we perform experiments which conclude a simple U-Net segmentation baseline without training tricks or complicated architectural changes is still a top performer for the task of change detection. All experiments are openly available at \href{https://github.com/isaaccorley/a-change-detection-reality-check}{github.com/isaaccorley/a-change-detection-reality-check}.
\end{abstract}

\section{Introduction}
\label{sec:intro}

The task of change detection from remotely sensed imagery is a canonical and important problem that allows us to analyze how our planet changes over time. It is crucial that change detection methods are highly accurate given their primary applications for building damage assessment disaster response~\citep{gupta2019creating,sublime2019automatic} and monitoring of the environment~\citep{cambrin2024cabuar,watch2002global}. The machine learning and remote sensing fields are currently experiencing their ImageNet~\citep{deng2009imagenet} benchmarking era with new proposed model architectures claiming incremental performance gains on standard benchmark datasets at a rapid rate. For example, the Changer paper~\citep{fang2023changer} -- first preprint released on Sep. 2022 -- is cited by 56 papers, the majority of them proposing a new architecture for change detection while also not releasing any open-source code or model checkpoints~\citep{ren2024mfinet,zhang2024mfi,zhou2024multistage,li4779358using,liu2024mutsimnet,huang2024remote,lin2024diformer,wang2024msgfnet,lu2024bi,tan2024bd,xu2024dual,yan2024multilevel,peng2023fda,wan2023cldrnet,zhao2023adapting,liu2023feature,quan2023unified,wang2023spatial,liu2023swin,fazry2023change}. Many of these proposed methods contain complicated architectural layers and modules specifically designed to better handle the bi-temporal image change detection task~\citep{bai2023deep,zhang2024bifa, yuan2024dynamically,li2024decoder,pang2024hicd,zheng2024detail,zhao2023gampf,li2023ms,yan2023vectorization,chen2023remote,liu2023attention,zhao2023gesanet}. Notably many of these papers are peer-reviewed and published at reputable conferences and journals. However, this begs the question if these new methods are actually improvements over generic segmentation architectures or have the benchmarks just been poorly executed? In this paper, we seek to answer this question.

\paragraph{Background}
When proposing a new model architecture it is common to perform a comparison to prior works through benchmark experiments to show statistically significant improvement. However over time, comparisons become less fair due to differences in training methodologies that are misconstrued as achieving state-of-the-art performance. Put simply, baseline experiments are often underpowered. This has been shown to be prevalent in many areas of research such as deep metric learning~\citep{musgrave2020metric}, unsupervised domain adaption~\citep{musgrave2021unsupervised}, image classification~\citep{bello2021revisiting}, deep reinforcement learning~\citep{henderson2018deep}, point cloud classification~\citep{uy2019revisiting}, and video recognition~\citep{du2021revisiting}. Furthermore, these analyses commonly conclude that simple baselines outperform complicated and task-specific architectures. Most recently, \citet{gerard2024simple} discovered that a generic architecture like U-Net, without training or evaluation tricks, is still competitive on the xBD dataset from the xView2 challenge~\citep{gupta2019xbd}. %

When analyzing relevant change detection papers and source code for benchmarking we find that it is common for authors to just compare to metrics reported in prior literature rather than re-running experiments with the prior methods in a consistent training setup. Further, we find that recent work often changes multiple aspects of the experimental setup, including the training routine (optimization methods, learning rate schedule, etc.) and loss functions, beyond just the proposed model architecture. This is problematic as any observed improvements on the benchmark dataset could be due to any combination of the new architecture, training routine, or loss function. This has the potential to result in unfair comparisons, especially if the improvement in quantitative performance is very small.

\textit{In this paper we revisit bi-temporal change detection benchmarks with simple baselines to get an overview on progress}. Specifically, we experiment with a simple semantic segmentation U-Net architecture, and siamese network variants~\citep{koch2015siamese}, to explore how these baselines perform against the latest state-of-the-art change detection methods. We find that this baseline architecture, from 2015, is a top performer on change detection benchmark datasets.

\section{Methods}
\label{methods}

\paragraph{State-of-the-Art Models}
In our experiments, in addition to compiling results from many prior change detection works, we retrain and evaluate the following state-of-the-art methods for change detection: BIT~\citep{chen2021remote} is a transformer-based siamese network architecture which uses a shared convolutional backbone to extract image features and transformer encoder decoder networks to perform change detection. ChangeFormer~\citep{bandara2022transformer} is an end-to-end transformer-based siamese network architecture for change detection. TinyCD~\citep{codegoni2023tinycd} is a change detection architecture which uses an EfficientNet~\citep{tan2019efficientnet} backbone to extract convolutional features to feed to a custom attention-based decoder network.

\paragraph{Our Baseline Models}

\citep{daudt2018fully} proposed three fully-convolutional (FC) architectures for change detection. The first is Early Fusion (FC-EF) which is an encoder-decoder style architecture with the change detection image pair concatenated as input. The other methods contain shared siamese encoders which either concatenate intermediate feature maps (FC-Siam-Conc) or take the difference (FC-Siam-Diff). While these networks are similar to the U-Net architecture~\citep{ronneberger2015u}, the original implementation is customized and hardcoded to be lightweight in parameter count and thus are unable to take advantage of pretrained encoder backbones like ResNet~\citep{he2016deep}. These methods are commonly seen in change detection benchmarks being compared to methods which utilize pretrained ImageNet backbones.

In our experiments, we generalize these implementations to use the standard U-Net framework which is able to take advantage of numerous ImageNet pretrained backbones. To avoid confusion with the original implementations of~\citep{daudt2018fully}, we refer to these architectures in our experiments as simply \texttt{U-Net}, \texttt{U-Net SiamConc}, and \texttt{U-Net SiamDiff}. Specifically, we utilize the implementations of these models in the TorchGeo~\citep{stewart2022torchgeo} library for reproducibility. We perform benchmarks using the ResNet-50 and EfficientNet-B4~\citep{tan2019efficientnet} backbones.

\begin{table}[t!]
\centering
\caption{Comparison of state-of-the-art and change detection architectures to a U-Net baseline on the LEVIR-CD dataset. We report the test set precision, recall, and F1 metrics of the positive change class. For the baseline experiments we perform 10 runs while varying random the seed and report metrics from the highest performing run. All other metrics are taken from their respective papers. The top performing methods are highlighted in bold. Gray rows indicate our baseline U-Net and siamese encoder variants.\\}
\resizebox{0.90\textwidth}{!}{%
\begin{tabular}{ccccc}
\toprule

\textbf{Model} &
\textbf{Backbone} &
\textbf{Precision} &
\textbf{Recall} &
\textbf{F1} \\
\toprule

FC-EF~\citep{daudt2018fully}         & -               & 86.91 & 80.17 & 83.40 \\
FC-Siam-Conc~\citep{daudt2018fully}   & -               & 91.99 & 76.77 & 83.69 \\
FC-Siam-Diff~\citep{daudt2018fully}   & -               & 89.53 & 83.31 & 86.31 \\
DTCDSCN~\citep{liu2020building}        & SE-Resnet34     & 88.53 & 86.83 & 87.67 \\
STANet~\citep{chen2020spatial}         & ResNet-18       & 83.81 & \textbf{91.00} & 87.26 \\
CDNet~\citep{chen2021adversarial}          & ResNet-18       & 91.60 & 86.50 & 89.00 \\
BIT~\citep{chen2021remote}            & ResNet-18       & 89.24 & 89.37 & 89.31 \\
ChangeFormer~\citep{bandara2022transformer}   & MiT-b1          & 92.59 & 89.68 & 91.11 \\
Tiny-CD~\citep{codegoni2023tinycd}        & EfficientNet-b4 & 92.68 & 89.47 & 91.05 \\
ChangerVanilla~\citep{fang2023changer} & ResNet-18       & 92.66 & 89.60 & 91.10 \\
ChangerEx~\citep{fang2023changer}      & ResNet-18       & 92.97 & 90.61 & \textbf{91.77} \\

\midrule
\midrule

\rowcolor{Gray}
\texttt{U-Net}~\citep{ronneberger2015u}          & EfficientNet-b4 & 92.69 & 87.16 & 89.25 \\
\rowcolor{Gray}
\texttt{U-Net}~\citep{ronneberger2015u}          & ResNet-50       & 91.97 & 89.78 & 90.38 \\
\rowcolor{Gray}
\texttt{U-Net SiamConc} & ResNet-50       & 92.87 & 89.48 & 90.41 \\
\rowcolor{Gray}
\texttt{U-Net SiamDiff} & ResNet-50       & \textbf{93.21} & 89.50 & 90.46 \\

\bottomrule
\end{tabular}%
}
\label{levircd-results}
\end{table}

\subsection{Datasets}
For comparisons of change detection architectures we utilize the following benchmark datasets:

\begin{description}
    \item[LEVIR-CD] \citep{chen2020spatial} A binary change detection dataset containing 637 high resolution (0.5m) 1024 × 1024 image pairs extracted from Google Earth. We utilize the splits provided with the dataset and, following other change detection papers, we convert the images to non-overlapping 256 × 256 patches.

    \item[WHU-CD] \citep{ji2018fully} A binary change detection dataset containing one pair of high-resolution (0.075m) aerial images of size 32507 × 15354. We utilize the train and test splits provided with the dataset and convert the images to non-overlapping 256 × 256 patches. During training runs we randomly split a 10\% holdout set from the train set to use as a validation set.
    
\end{description}

\section{Experiments}
\label{experiments}

\subsection{Baseline Training Details}
\label{sec:baseline-training-details}

Throughout our baseline training experiments we use the same hyperparameter setup from the BIT~\citep{chen2021remote} source which consists of batch size of 8, 200 epochs, stochastic gradient descent (SGD) optimizer with an initial learning rate $\gamma=0.01$, momentum $\mu=0.9$, weight decay $\alpha=5E-4$, and a linear decaying scheduler. We train each model by optimizing a cross entropy loss on a multiclass segmentation output where the number of classes is 2 for binary change detection. We select the checkpoint with the top performance based on the validation loss and do not perform any early stopping. We note that this setup results in a number of iterations which allows each model to converge.

During training we use augmentations consisting of random horizontal and vertical flips with probability $p=0.5$ and random resize crop with scale in the range $[0.8, 1.0]$, aspect ratio of $1$, and $p=1$. For image normalization we simply rescale the images to the range $[-1, 1]$ following the BIT~\citep{chen2021remote} source.

\subsection{Updated WHU-CD Benchmarks}
Many papers utilize a preprocessed and randomly split version of the WHU-CD dataset created in \citep{bandara2022revisiting} for change detection benchmarks. However these sets have been known to introduce data leakage~\citep{wgcbanchangeformergithubissue1} -- $\approx85\%$ of the test set is included in the train set due to a bug in the preprocessing scripts -- which makes them impossible to use to benchmark methods. We perform a new benchmark using the original train and test set splits from the WHU-CD dataset and retrain the BIT, ChangeFormer, and TinyCD models for comparison. We use the same experimental settings described in Section~\ref{sec:baseline-training-details}.

We find that performance of some models, particularly the transformer based methods BIT and ChangeFormer, can vary significantly over different random seeds, therefore we train each model on each dataset for runs with 10 different seeds. For comparisons we select the results from the highest performing run as well as report the average and standard deviation over runs for transparency. Any additional parameters can be found in our open source implementation.

\begin{table}[ht!]
\centering
\caption{Experimental results on the WHU-CD dataset. We retrain several state-of-the-art methods using the original dataset's train/test splits instead of the commonly used randomly split preprocessed version created in ~\cite{bandara2022revisiting}. We find that these state-of-the-art methods are outperformed by a U-Net baseline. We report the test set precision, recall, F1, and IoU metrics of the positive change class. For each run we select the model checkpoint with the lowest validation set loss. We provide metrics averaged over 10 runs with varying random seed as well as the best seed. Gray rows indicate our baseline U-Net and siamese encoder variants.\\}
\resizebox{0.90\textwidth}{!}{%
\begin{tabular}{cccccc}
\toprule
\textbf{Model} &
\textbf{Backbone} &
\textbf{F1} &
\textbf{Pre.} &
\textbf{Rec.} &
\textbf{IoU} \\
\toprule

\multicolumn{6}{l}{\textbf{Averaged Over 10 Seeds}} \\
\midrule
ChangeFormer & MiT-b1 & 75.65 ± 1.58 & 77.06 ± 3.22 & 74.67 ± 1.97 & 61.60 ± 2.05 \\
TinyCD & EfficientNet-b4 & 78.53 ± 1.28 & 80.15 ± 2.49 & 77.56 ± 2.13 & 65.52 ± 1.72 \\
BIT & ResNet-18 & 72.67 ± 2.69 & 70.30 ± 6.36 & 76.84 ± 4.53 & 58.06 ± 3.24 \\
\rowcolor{Gray}
\texttt{U-Net} & ResNet-50 & 81.85 ± 1.32 & 83.72 ± 2.65 & 80.39 ± 2.32 & 69.96 ± 1.83 \\
\rowcolor{Gray}
\texttt{U-Net SiamConc} & ResNet-50 & 81.33 ± 1.08 & 79.30 ± 2.78 & \textbf{84.19 ± 1.55} & 69.40 ± 1.52 \\
\rowcolor{Gray}
\texttt{U-Net SiamDiff} & ResNet-50 & \textbf{82.02 ± 1.48} & \textbf{83.82 ± 3.80} & 80.92 ± 2.51 & \textbf{70.29 ± 2.00} \\

\midrule
\multicolumn{6}{l}{\textbf{Best Seed}} \\
\midrule
ChangeFormer   & MiT-b1          & 77.75          & 82.60           & 78.57          & 64.22          \\
TinyCD         & EfficientNet-b4 & 78.53          & 80.15          & 77.56          & 65.52          \\
BIT            & ResNet-18       & 77.68          & 78.58          & 82.13          & 64.34          \\
\rowcolor{Gray}
\texttt{U-Net}          & ResNet-50       & \textbf{84.17} & \textbf{88.65} & 83.08          & \textbf{73.23} \\
\rowcolor{Gray}
\texttt{U-Net SiamConc} & ResNet-50       & 82.75          & 83.69          & \textbf{86.56} & 71.15          \\
\rowcolor{Gray}
\texttt{U-Net SiamDiff} & ResNet-50       & 84.01          & 88.56          & 85.63          & 73.02         \\

\bottomrule
\end{tabular}%
}
\label{whucd-results}
\end{table}

\section{Discussion}

\paragraph{Results}
In Table~\ref{levircd-results} we provide experimental results for our baseline on the LEVIR-CD dataset. We compare to prior work by compiling results from their respective papers. While the compiled results represent the best reported metrics for each previous method, it still is evident that a simple U-Net baseline is a top performer. Table~\ref{whucd-results} provides additional updated experimental results for the WHU-CD dataset and compares several state-of-the-art change detection architectures. Again, we can see that the baselines outperform all other methods.

Regarding the siamese variants of our U-Net baseline (\texttt{U-Net SiamDiff} and \texttt{U-Net SiamConc}), our experimental results indicate that processing each image in the bi-temporal pair with a shared encoder results in reasonable gains and is a promising approach. Methods such as Changer~\citep{fang2023changer} also conclude that additional feature interaction is key for achieving performance improvements for change detection.

\paragraph{Limitations}
Due to the fast pace of the field of change detection, with new architectures being proposed almost weekly, it is difficult to benchmark each new method. Furthermore, we do not compare to methods which propose a technique, e.g. a loss function, which is dependent on the architecture or methods, or to methods which do not publish open source code to reproduce the experimental results.

We note that there exists several other common change detection datasets such as DSIFN-CD~\citep{zhang2020deeply}, S2Looking~\citep{shen2021s2looking}, SECOND~\citep{yang2020semantic}, LEVIR-CD+~\citep{shen2021s2looking}, and xBD~\citep{gupta2019xbd}. While we do not benchmark against these, we leave this for future work.

\section{Conclusion and Next Steps}
In this paper we analyzed whether the field of change detection has actually made significant improvements on benchmark datasets in recent years. We conclude that many claimed improvements are questionable by demonstrating that a simple baseline of U-Net is still a top-performing method. To be clear, this is not an issue unique to change detection; other machine learning fields such as language modeling are finding it crucial to standardize fair benchmarking~\citep{srivastava2022beyond} when new methods are rapidly proposed. To mitigate this, we recommend utilizing and contributing proposed models to libraries and projects such as OpenCD~\citep{opencd}, GEO-Bench~\citep{lacoste2023geo}, and TorchGeo~\citep{stewart2022torchgeo} which standardize datasets and trainers for reliable benchmarking of remote sensing tasks. As this field has important downstream applications, we hope our results motivate the community to perform more reliable benchmarks of performance so that realistic advancements in change detection can be achieved. 

\subsubsection*{Acknowledgments}
We thank Jonathan Lwowski, Conor Wallace, Robin Cole, Sebastian Gerard, and Juan M. Lavista Ferres for their valuable feedback.

\bibliography{iclr2024_conference}
\bibliographystyle{iclr2024_conference}

\end{document}